\documentclass{article}

\pdfoutput=1

\usepackage[final]{nips_2017}

\usepackage[numbers]{natbib}

\usepackage[utf8]{inputenc}
\usepackage{natbib}
\usepackage{graphicx}
\usepackage{xcolor}
\usepackage{subcaption}
\usepackage{hyperref}
\usepackage{multirow}
\usepackage{siunitx}
\usepackage{amssymb}

\definecolor{Maroon}{cmyk}{0, 0.87, 0.68, 0.32}

\newcommand{\footref}[2]{\href{#1}{#2}\footnote{\url{#1}}}

\def\E#1{\mathbb{E}\left[#1\right]}

\title{StarCraft II: A New Challenge for\\ Reinforcement Learning}

\date{August 2017}

\begin{document}

\frenchspacing % No extra space after periods.

\maketitle

{\bf\center{
Oriol~Vinyals \hskip 0.2in Timo~Ewalds \hskip 0.2in Sergey~Bartunov \hskip 0.2in Petko~Georgiev \hskip 0.2in Alexander~Sasha~Vezhnevets \hskip 0.2in Michelle~Yeo \hskip 0.2in Alireza~Makhzani \hskip 0.2in Heinrich~Küttler \hskip 0.2in John~Agapiou \hskip 0.2in Julian~Schrittwieser \hskip  0.2in John Quan \hskip 0.2in Stephen~Gaffney \hskip 0.2in Stig~Petersen \hskip 0.2in Karen~Simonyan \hskip 0.2in Tom~Schaul \hskip 0.2in Hado~van~Hasselt \hskip 0.2in David~Silver \hskip 0.2in Timothy~Lillicrap \\
\emph{DeepMind} \\ \vskip 0.2in
Kevin~Calderone \hskip 0.2in Paul~Keet \hskip 0.2in Anthony~Brunasso \hskip 0.2in David~Lawrence \hskip 0.2in Anders~Ekermo \hskip 0.2in Jacob~Repp \hskip 0.2in Rodney~Tsing \\
\emph{Blizzard} \\
}}

\vskip 0.3in
\begin{abstract}
This paper introduces \emph{SC2LE} (StarCraft II Learning Environment), a reinforcement learning environment based on the game StarCraft II. This domain poses a new grand challenge for reinforcement learning, representing a more difficult class of problems than considered in most prior work. It is a multi-agent problem with multiple players interacting; there is imperfect information due to a partially observed map; it has a large action space involving the selection and control of hundreds of units; it has a large state space that must be observed solely from raw input feature planes; and it has delayed credit assignment requiring long-term strategies over thousands of steps. We describe the observation, action, and reward specification for the StarCraft II domain and provide an open source Python-based interface for communicating with the game engine. In addition to the main game maps, we provide a suite of mini-games focusing on different elements of StarCraft II gameplay.
For the main game maps, we also provide an accompanying dataset of game replay data from human expert players. We give initial baseline results for neural networks trained from this data to predict game outcomes and player actions.
Finally, we present initial baseline results for canonical deep reinforcement learning agents applied to the StarCraft II domain. On the mini-games, these agents learn to achieve a level of play that is comparable to a novice player. However, when trained on the main game, these agents are unable to make significant progress. Thus, \emph{SC2LE} offers a new and challenging environment for exploring deep reinforcement learning algorithms and architectures.
\end{abstract}

\section{Introduction}

Recent progress in areas such as speech recognition \cite{Dahl:2012}, computer vision \cite{Krizhevsky:2012}, and natural language processing \cite{Wu:2016google} can be attributed to the resurgence of deep learning \cite{Lecun:2015deep}, which provides a powerful toolkit for non-linear function approximation using neural networks. These techniques have also proven successful in reinforcement learning problems, yielding significant successes in Atari \cite{mnih2015human}, the game of Go \cite{silver2016mastering}, three-dimensional virtual environments \cite{Beattie:2016} and simulated robotics domains \cite{Levine:2016,Rusu:2016}. Many of the successes have been stimulated by the availability of simulated domains with an appropriate level of difficulty. Benchmarks have been critical to measuring and therefore advancing deep learning and reinforcement learning (RL) research \cite{bellemare2013arcade,mnih2015human,DBLP:journals/ijcv/RussakovskyDSKS15,Duan:2016}. It is therefore important to ensure the availability of domains that are \emph{beyond} the capabilities of current methods in one or more dimensions.

In this paper we introduce \emph{SC2LE}\footnote{Pronounced: ``school''.} (StarCraft II Learning Environment), a challenging domain for reinforcement learning, based on the StarCraft II video game. StarCraft is a real-time strategy (RTS) game that combines fast paced micro-actions with the need for high-level planning and execution. Over the previous two decades, StarCraft I and II have been pioneering and enduring e-sports\footref{https://en.wikipedia.org/wiki/Professional_StarCraft_competition}, with millions of casual and highly competitive professional players. Defeating top human players therefore becomes a meaningful and measurable long-term objective. 

From a reinforcement learning perspective, StarCraft II also offers an unparalleled opportunity to explore many challenging new frontiers. First, it is a multi-agent problem in which several players compete for influence and resources. It is also multi-agent at a lower-level: each player controls hundreds of units, which need to collaborate to achieve a common goal. Second, it is an imperfect information game. The map is only partially observed via a local camera, which must be actively moved in order for the player to integrate information. Furthermore, there is a ``fog-of-war'', obscuring the unvisited regions of the map, and it is necessary to actively explore the map in order to determine the opponent's state. Third, the action space is vast and diverse. The player selects actions among a combinatorial space of approximately $10^8$ possibilities (depending on the game resolution), using a point-and-click interface. There are many different unit and building types, each with unique local actions. Furthermore, the set of legal actions varies as the player progresses through a tree of possible technologies. Fourth, games typically last for many thousands of frames and actions, and the player must make early decisions (such as which units to build) with consequences that may not be seen until much later in the game (when the players' armies meet), leading to a rich set of challenges in temporal credit assignment and exploration.

This paper introduces an interface intended to make RL in StarCraft straightforward: observations and actions are defined in terms of low resolution grids of features; rewards are based on the score from the StarCraft II engine against the built-in computer opponent; and several simplified mini-games are also provided in addition to the full game maps. Future releases will extend the interface for the full challenge of StarCraft II: observations and actions will expose RGB pixels; agents will be ranked by the final win/loss outcome in multi-player games; and evaluation will be restricted to full game maps used in competitive human play.

In addition, we provide a large dataset based on game replays recorded from human players, which will increase to millions of replays as people play the game. We believe that the combination of the interface and this dataset will provide a useful benchmark to test not only existing and new RL algorithms, but also interesting aspects of perception, memory and attention, sequence prediction, and modelling uncertainty, all of which are active areas of machine learning research.

Several environments \cite{BWAPI, Tian:2017elf, synnaeve2016torchcraft} already exist for reinforcement learning in the original version of StarCraft. Our work differs from these previous environments in several regards: it focuses on the newer version StarCraft II; observations and actions are based on the human user interface rather than being programmatic; and it is directly supported by the game developers, Blizzard Entertainment, on Windows, Mac, and Linux.

The current best artificial StarCraft bots, based on the built-in AI or research on previous environments, can be defeated by even amateur players \cite[cf.][and later versions of the AIIDE competition]{Buro:2012real}. This fact, coupled with StarCraft's interesting set of game-play properties and large player base, makes it an ideal research environment for exploring deep reinforcement learning algorithms.

\section{Related Work}

Computer games provide a compelling solution to the issue of evaluating and comparing different learning and planning approaches on standardised tasks, and is an important source of challenges for research in artificial intelligence (AI).  These games offer multiple advantages: 1.\ They have clear objective measures of success; 2.\ Computer games typically output rich streams of observational data, which are ideal inputs for deep networks; 3.\ They are externally defined to be difficult and interesting for a human to play.  This ensures that the challenge itself is not tuned by the researcher to make the problem easier for the algorithms being developed; 4.\ Games are designed to be run anywhere with the same interface and game dynamics, making it easy to share a challenge precisely with other researchers;  5.\ In some cases a pool of avid human players exists, making it possible to benchmark against highly skilled individuals. 6.\ Since games are simulations, they can be controlled precisely, and run at scale.

A well known example of games driving reinforcement learning research is the Arcade Learning Environment (ALE \cite{bellemare2013arcade}), which allows easy and replicable experiments with Atari video games.  This standardised set of tasks has been an incredible boon to recent research in AI.  Scores on games in this environment can be compared across publications and algorithms, allowing for direct measurement and comparison.
The ALE is a prominent example in a rich tradition of video game benchmarks for AI~\cite{measuring:2011}, including Super Mario \cite{supermario}, Ms Pac-Man \cite{rohlfshagen2011ms}, Doom~\cite{vizdoom}, Unreal Tournament \cite{hingston2009turing}, as well as general video game-playing frameworks \cite{schaul2013video,bhonker2016playing} and competitions \cite{perez20152014}.

The genre of RTS games has attracted a large amount of AI research, including on the original StarCraft (Broodwar). We recommend the surveys by Ontanon et al.~\cite{ontanon2013survey} and  Robertson \& Watson \cite{robertson2014review} for an overview. Many of those research directions focus on specific aspects of the game (e.g., build order, or combat micro-management) or specific AI techniques (e.g., MCTS planning). We are not aware of efforts to solve full games with an end-to-end RL approach. Tackling full versions of RTS games has seemed daunting because of the rich input and output spaces as well as the very sparse reward structure (i.e., game outcome).

The standard API for StarCraft thus far has been BWAPI \cite{BWAPI}, and related wrappers   \cite{synnaeve2016torchcraft}. Simplified versions of RTS games have also been developed for AI research, most notably \footref{https://github.com/santiontanon/microrts}{microRTS} or the more recent ELF \cite{tian2017elf}. Previous work has applied RL approaches to the Wargus RTS game with reduced state and action spaces \cite{jaidee2012classq}, and learning based agents have also been explored in micro-management mini-games \cite{peng2017multiagent, usunier2016episodic}, and learning game outcome or \emph{build orders} from replay data \cite{erickson2014global,justesen2017learning}.
\section{The SC2LE Environment}\label{sec:env}
\begin{figure}
    \centering
    \includegraphics[width=\linewidth]{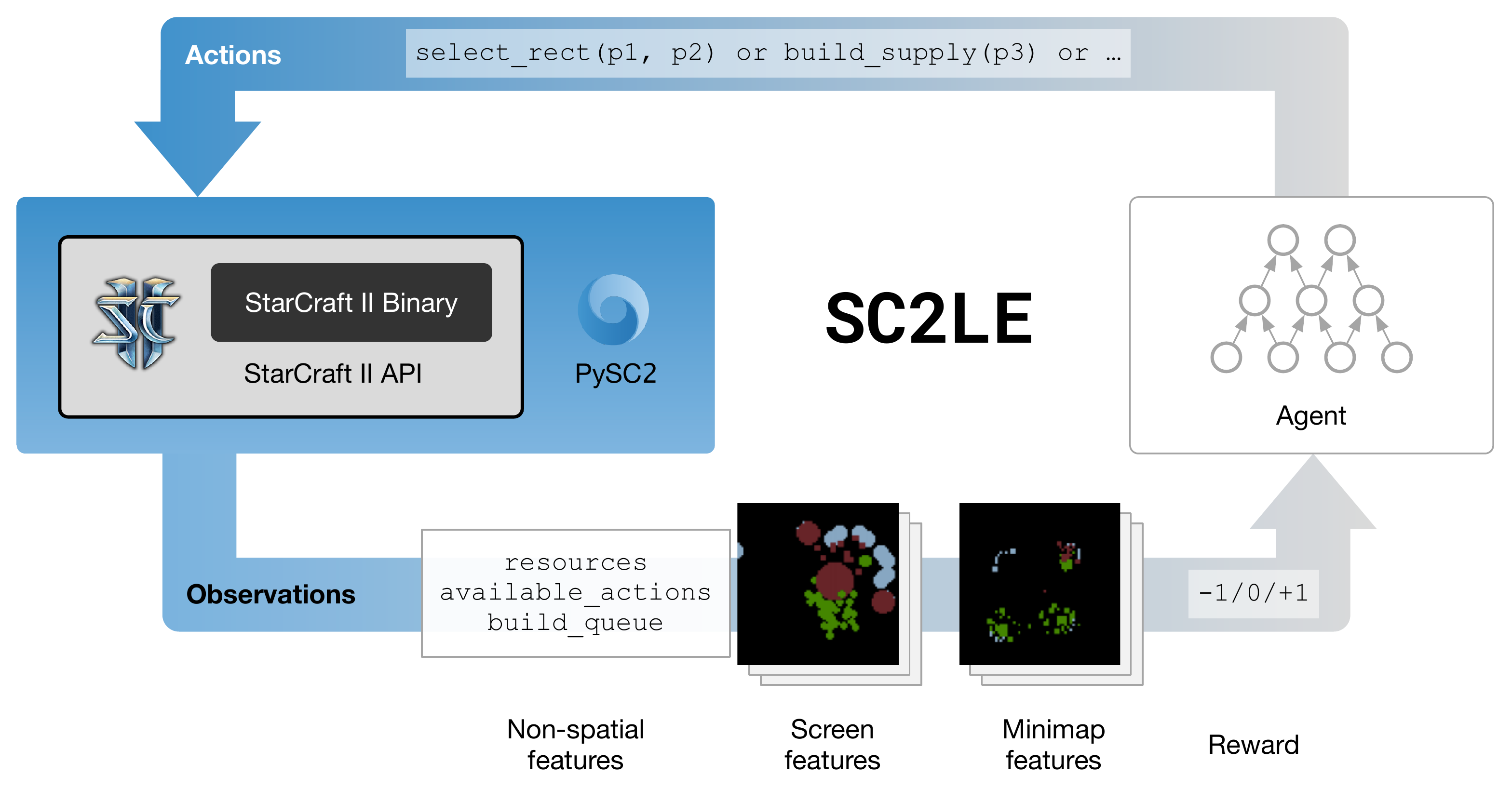}
    \caption{The StarCraft II Learning Environment, \emph{SC2LE}, shown with its components plugged into a neural agent.
    }
    \label{fig:framework}
\end{figure}

The main contribution of our paper is the release of \emph{SC2LE}, which exposes StarCraft II as a research environment. The release consists of three sub-components: a Linux StarCraft II binary, the StarCraft~II API, and PySC2 (see figure~\ref{fig:framework}).

The \footref{https://github.com/Blizzard/s2client-proto}{StarCraft II API} allows programmatic control of StarCraft II. The API can be used to start a game, get observations, take actions, and review replays. This API into the normal game is available on Windows and Mac OS, but we also provide a limited headless build that runs on Linux especially for machine learning and distributed use cases.

Using this API we built \footref{https://github.com/deepmind/pysc2}{PySC2}, an open source environment that is optimised for RL agents. PySC2 is a Python environment that wraps the StarCraft II API to ease the interaction between Python reinforcement learning agents and StarCraft II. PySC2 defines an action and observation specification, and includes a random agent and a handful of rule-based agents as examples. It also includes some mini-games as challenges and visualisation tools to understand what the agent can see and do.

StarCraft II updates the simulation 16 (at ``normal speed'') or 22.4 (at ``fast speed'') times per second.
The game is mostly deterministic, but it does have some randomness mainly for cosmetic reasons; the two main random elements are weapon speed and update order. These sources of randomness can be removed/mitigated by setting a random seed.

We now describe the environment which was used for all of the experiments in this paper.

\subsection{Full Game Description and Reward Structure}

In the full 1v1 game of StarCraft II, two opponents spawn on a map which contains resources and other elements such as ramps, bottlenecks, and islands. To win a game, a player must: 1.\ Accumulate resources (minerals and vespene gas), 2.\ Construct production buildings, 3.\ Amass an army, and 4.\ Eliminate all of the opponent's buildings. A game typically lasts from a few minutes to one hour, and early actions taken in the game (e.g., which buildings and units are built) have long term consequences. Players have imperfect information since they can typically only see the portion of the map where they have units. If they want to understand and react to their opponent's strategy they must send units to scout. As we describe later in this section, the action space is also quite unique and challenging.

Most people play online against other human players. The most common games are 1v1, but team games are possible too (2v2, 3v3 or 4v4), as are more complicated games with unbalanced teams or more than two teams. Here we focus on the 1v1 format, the most popular form of competitive StarCraft, but may consider more complicated situations in the future.

StarCraft II includes a built-in AI which is based on a set of handcrafted rules and comes with 10~levels of difficulty (the three strongest of which cheat by getting extra resources or privileged vision). Unfortunately, the fact that they are rule-based means their strategies are fairly narrow and thus easily exploitable. Nevertheless, they are a reasonable first challenge for a purely learned approach like the baselines we investigate in sections~\ref{sec:rl} and \ref{sec:sl}; they play far better than random, play very quickly with little compute, and offer consistent baselines to compare against. 

We define two different reward structures: ternary $1$~(win) / $0$~(tie) / $-1$~(loss) received at the end of a game (with all-zero rewards during the game), and Blizzard score. The ternary win/tie/loss score is the real reward that we care about. The Blizzard score is the score seen by players on the victory screen at the end of the game. While players can only see this score at the end of the game, we provide access to the running Blizzard score at every step during the game so that the change in score can be used as a reward for reinforcement learning. It is computed as the sum of current resources and upgrades researched, as well as units and buildings currently alive and being built. This means that the player's cumulative reward increases with more mined resources, decreases when losing units/buildings, and all other actions (training units, building buildings, and researching) do not affect it. The Blizzard score is not zero-sum since it is player-centric, it is far less sparse than the ternary reward signal, and it correlates to some extent with winning or losing.

\subsection{Observations}

\begin{figure}
    \centering
    \includegraphics[width=\linewidth]{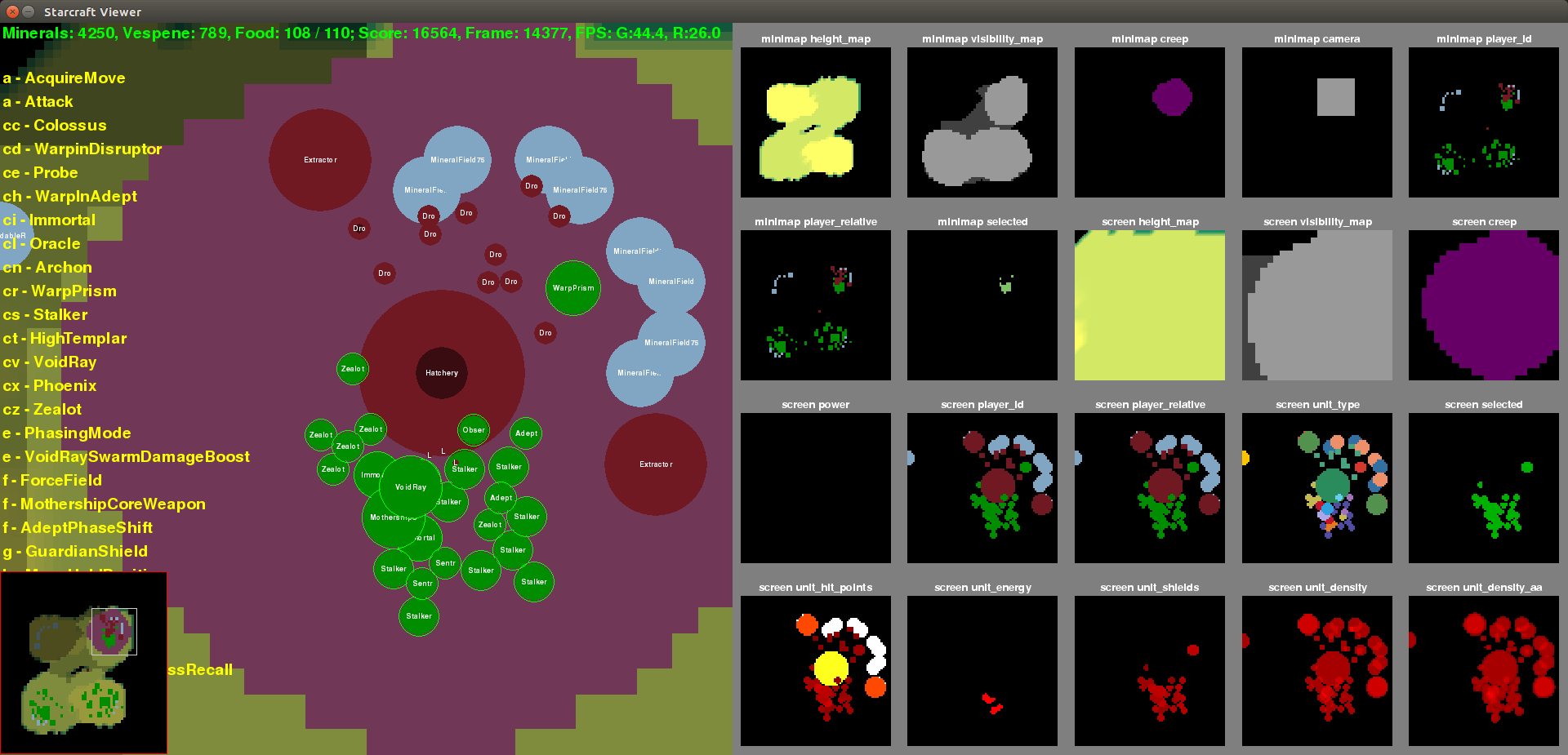}
    \caption{The PySC2 viewer shows a human interpretable view of the game on the left, and coloured versions of the feature layers on the right.  For example, terrain height, fog-of-war, creep, camera location, and player identity, are shown in the top row of feature layers. A video can be found at \url{https://youtu.be/-fKUyT14G-8}.
    }
    \label{fig:viewer}
\end{figure}

StarCraft II uses a game engine which renders graphics in 3D. Whilst utilising the underlying game engine which simulates the whole environment, the StarCraft II API does not currently render RGB pixels. Rather, it generates a set of ``feature layers'', which abstract away from the RGB images seen during human play, while maintaining the core spatial and graphical concepts of StarCraft II (see Figure~\ref{fig:viewer}). 

Thus, the main observations come as sets of feature layers which are rendered at $N\times M$ pixels (where $N$ and $M$ are configurable, though in our experiments we always used $N=M$). Each of these layers represents something specific in the game, for example: unit type, hit points, owner, or visibility. Some of these (e.g., hit points, height map) are scalars, while others (e.g., visibility, unit type, owner) are categorical. There are two sets of feature layers: the \emph{minimap} is a coarse representation of the state of the entire world, and the \emph{screen} is a detailed view of a subsection of the world corresponding to the player's on-screen view, and in which most actions are executed. Some features (e.g., owner or visibility) exist for both the screen and minimap, while others (e.g., unit type and hit points) exist only on the screen. See the \footref{https://github.com/deepmind/pysc2/blob/master/docs/environment.md}{environment documentation} for a complete description of all observations provided.

In addition to the screen and minimap, the human interface for the game provides various non-spatial observations. These include the amount of gas and minerals collected, the set of actions currently available (which depends on game context, e.g., which units are selected), detailed information about selected units, build queues, and units in a transport vehicle. These observations are also exposed by PySC2, and are fully described in the \href{https://github.com/deepmind/pysc2/blob/master/docs/environment.md}{environment documentation}. The audio channel is not exposed as a wave form but important notifications will be exposed as part of the observations.

In the retail game engine the screen is rendered with a full 3D perspective camera at high resolution. This leads to complicated observations with units getting smaller as they get ``higher'' on the screen, and with more world real estate being visible in the back than the front. To simplify this, feature layers are rendered via a camera that uses a top down orthographic projection. This means that each pixel in a feature layer corresponds to precisely the same amount of world real estate, and as a consequence all units will be the same size regardless where they are in view. Unfortunately, it also means the feature layer rendering does not quite match what a human would see. An agent sees a little more in the front and a little less in the back. This does mean some actions that humans make in replays cannot be fully represented.

In future releases we will expose a rendered API allowing agents to play from RGB pixels. This will allow us to study the effects of learning from raw pixels versus learning from feature layers and make closer comparisons to human play. In the mean time, we played the game with feature layers to verify that agents are not severely handicapped. Though the game-play experience is obviously altered we found that a resolution of $N, M \ge 64$ is sufficient to allow a human player to select and individually control small units such as Zerglings. The reader is encouraged to try this using \footref{https://github.com/deepmind/pysc2/blob/master/pysc2/bin/play.py}{\tt pysc2\_play}. See also Figure~\ref{fig:viewer}.

\subsection{Actions}

\begin{figure}
    \centering
    \includegraphics[width=\linewidth]{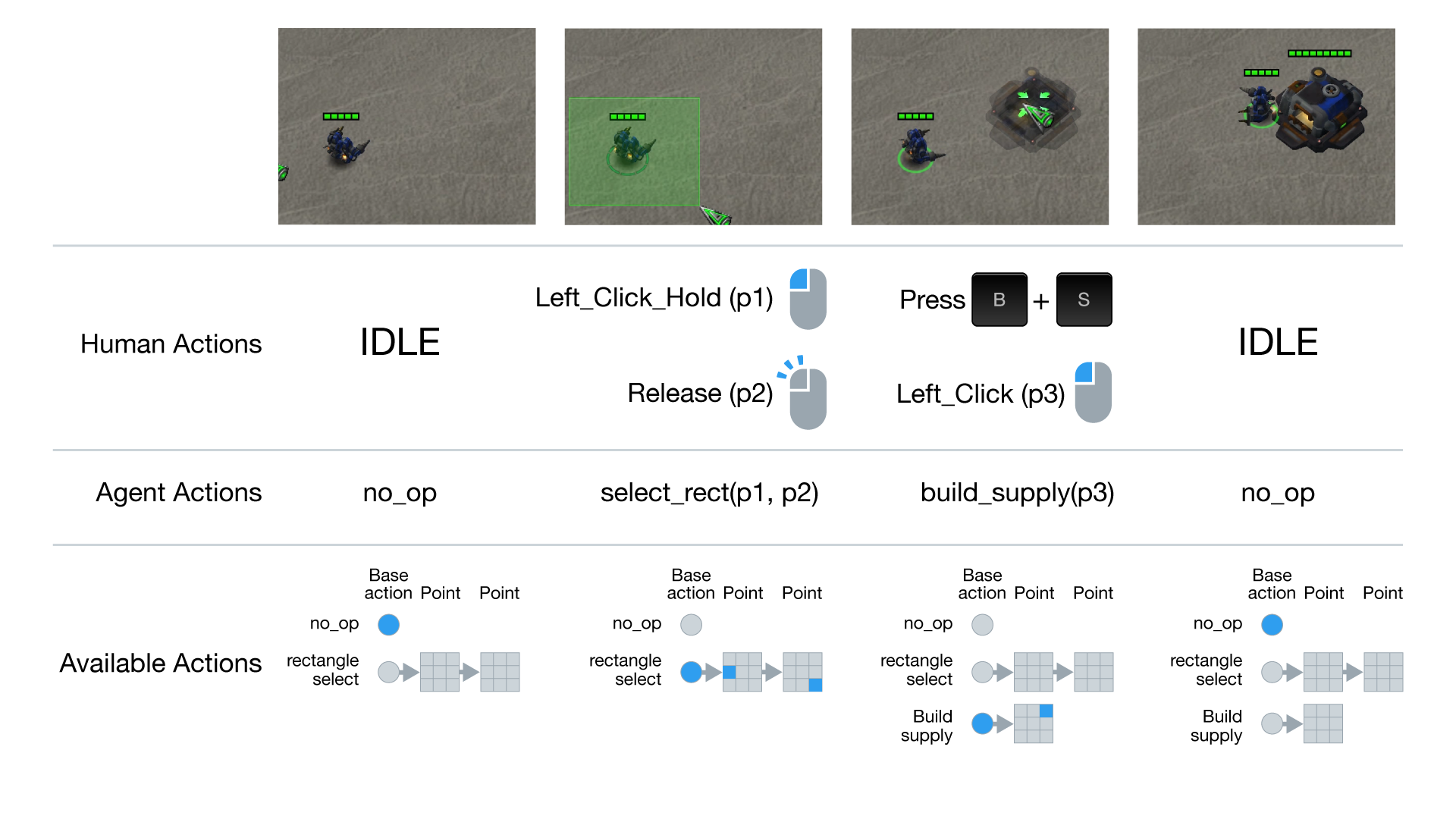}
    \caption{Comparison between how humans act on StarCraft II and the actions exposed by PySC2. We designed the action space to be as close as possible to human actions. The first row shows the game screen, the second row the human actions, the third row the logical action taken in PySC2, and the fourth row the actions $a$ exposed by the environment (and, in red, what the agent selected at each time step). Note that the first two columns do not feature the `build supply' action, as it is not yet available to the agent in those situations as a worker has to be selected first.
    }
    \label{fig:actions}
\end{figure}

We designed the environment action space to mimic the human interface as closely as possible whilst maintaining some of the conventions employed in other RL environments, such as Atari \cite{bellemare2013arcade}. Figure~\ref{fig:actions} shows a short sequence of actions as produced by a player and by an agent.

Many basic manoeuvres in the game are compound actions. For example, to move a selected unit across the map a player must first choose to move it by pressing \textit{m}, then possibly choose to queue the action by holding \textit{shift}, then click a point on the screen or minimap to execute the action. Instead of asking agents to produce those 3 key/mouse presses as a sequence of three separate actions we give it as an atomic compound function action: {\tt move\_screen(queued, screen)}.

More formally, an action $a$ is represented as a composition of a function identifier $a^0$ and a sequence of arguments which that function identifier requires: $a^1, a^2, \ldots, a^L$. For instance, consider selecting multiple units by drawing a rectangle. The intended action is then $\texttt{select\_rect}(\texttt{select\_add}, (x^1, y^1), (x^2, y^2))$.  The first argument $\texttt{select\_add}$ is binary.  The other arguments are integers that define coordinates --- their allowed range is the same as the resolution of the observations.
This action is fed to the environment in the form $[\texttt{select\_rect}, [[\texttt{select\_add}], [x^1, y^1], [x^2, y^2]]]$.

To represent the full action space we define approximately 300 action-function identifiers with 13 possible types of arguments (ranging from binary to specifying a point on the discretised 2D screen).
See the \href{https://github.com/deepmind/pysc2/blob/master/docs/environment.md}{environment documentation} for a more detailed specification and description of the actions available through PySC2, and Figure~\ref{fig:actions} for an example of a sequence of actions.

In StarCraft, not all the actions are available in every game state. For example, the move command is only available if a unit is selected. Human players can see which actions are available in the ``command card'' on the screen. Similarly, we provide a list of available actions via the observations given to the agent at each step. Taking an action that is not available is considered an error, so agents should filter their action choices so that only legal actions are taken.

Humans typically make between 30 and 300 actions per minute (APM), roughly increasing with player skill, with professional players often spiking above 500 APM. In all our RL experiments, we act every 8 game frames, equivalent to about 180 APM, which is a reasonable choice for intermediate players.

We believe these early design choices make our environment a promising testbed for developing complex RL agents.  In particular, the fixed-size feature layer input space and human-like action space are natural for neural network based agents.  This is in contrast to other recent work \cite{synnaeve2016torchcraft, peng2017multiagent}, where the game is accessed on a unit-per-unit basis and actions are individually specified to each unit.  While there are advantages to both interface styles, PySC2 offers the following:

\begin{itemize}
    \item Learning from human replays becomes simpler.
    \item We do not require unrealistic/super-human actions per minute to issue instructions individually to each unit.
    \item The game was designed to be played with this UI, and the balance between strategic high level decisions, managing your economy, and controlling the army makes the game more interesting.
\end{itemize}

\subsection{Mini-Games Task Description}

To investigate elements of the game in isolation, and to provide further fine-grained steps towards playing the full game, we built several mini-games. These are focused scenarios on small maps that have been constructed with the purpose of testing a subset of actions and/or game mechanics with a clear reward structure. Unlike the full game where the reward is 
just win/lose/tie,
the reward structure for mini-games can reward particular behaviours (as defined in a corresponding \texttt{.SC2Map} file). 

We encourage the community to build modifications or new mini-games with the powerful StarCraft Map Editor. This allows for more than just designing a broad range of smaller challenge domains. It permits sharing identical setups and evaluations with other researchers and obtaining directly comparable evaluation scores. The restricted action sets, custom reward functions and/or time limits are defined directly in the resulting \texttt{.SC2Map} file, which is easy to share. We therefore encourage users to use this method of defining new tasks, rather than customising on the agent side.

The seven mini-games that we are releasing are as follows:

\begin{itemize}
\item MoveToBeacon: The agent has a single marine that gets $+1$ each time it reaches a beacon. This map is a unit test with a trivial greedy strategy.
\item CollectMineralShards: The agent starts with two marines and must select and move them to pick up mineral shards spread around the map. The more efficiently it moves the units, the higher the score.
\item FindAndDefeatZerglings: The agent starts with 3 marines and must explore a map to find and defeat individual Zerglings. This requires moving the camera and efficient exploration.
\item DefeatRoaches: The agent starts with 9 marines and must defeat 4 roaches. Every time it defeats all of the roaches it gets 5 more marines as reinforcements and 4 new roaches spawn. The reward is $+10$ per roach killed and $-1$ per marine killed. The more marines it can keep alive, the more roaches it can defeat.
\item DefeatZerglingsAndBanelings: The same as DefeatRoaches, except the opponent has Zerglings and Banelings, which give $+5$ reward each when killed. This requires a different strategy because the enemy units have different abilities.
\item CollectMineralsAndGas: The agent starts with a limited base and is rewarded for the total resources collected in a limited time. A successful agent must build more workers and expand to increase its resource collection rate.
\item BuildMarines: The agent starts with a limited base and is rewarded for building marines. It must build workers, collect resources, build Supply Depots, build Barracks, and then train marines. The action space is limited to the minimum action set needed to accomplish this goal.
\end{itemize}

All mini-games have a fixed time limit and are described in more detail online: \url{
https://github.com/deepmind/pysc2/blob/master/docs/mini_games.md}.

\subsection{Raw API}

StarCraft II also has a \emph{raw} API, which is similar to the Broodwar API (BWAPI \cite{BWAPI}). 
In this case, the observations are a list of all visible units on the map along with the properties (unit type, owner, coordinates, health, etc.), but without any visual component. Fog-of-war still exists, but there is no camera, so you can see all visible units simultaneously. This is a simpler and more precise representation, but it does not correspond to how humans perceive the game. For the purposes of comparing against humans this is considered ``cheating'' since it offers significant additional information.

Using the raw API, actions control units or groups of units individually by a unit identifier. There is no need to select individuals or groups of units before issuing actions. This allows much more precise actions than the human interface allows, and thus yields the possibility of super-human behaviour via this API.

Although we have not used any data from the raw API to train our agents, it is included in the release in order to support other use cases. PySC2 uses it for visualization while both \footref{https://github.com/Blizzard/s2client-api}{Blizzard's SC2 API examples} and \footref{https://github.com/davechurchill/CommandCenter}{CommandCenter} use it to for rule-based agents.

\subsection{Performance}

We can often run the environment faster than real time.
Observations are rendered at a speed that depends on several factors: the map complexity, the screen resolution, the number of non-rendered frames per action, and the number of threads.

For complex maps (e.g., full ladder maps) the computation is dominated by simulation speed. Taking actions less often, allowing for fewer rendered frames, reduces the compute, but diminishing returns kicks in fairly quickly meaning there is little gain above 8 steps per action. Given little time is spent rendering, a higher resolution does not hurt. Running more instances in parallel threads scales quite well.

For simpler maps (e.g., CollectMineralShards) the world simulation is quick, so rendering the observations dominates.  In this case increasing the frames per action and decreasing the resolution can have a large effect. The bottleneck then becomes the Python interpreter, negating gains above roughly 4 threads with a single interpreter.

With a resolution of $64\times 64$ and acting at a rate of 8 frames per action, the single-threaded speed of a ladder map varies from 200--700 game steps per wall-clock second, which is more than an order of magnitude faster than real-time.  The exact speeds depends on multiple factors, including: the stage of the game, the number of units in play, and the computer it runs on. On CollectMineralShards the same settings permit 1600--2000 game steps per wall-clock second.

\section{Reinforcement Learning: Baseline Agents}\label{sec:rl}

This section provides baseline results that serve to calibrate the map difficulty, and demonstrate that established RL algorithms can learn useful policies, at least on the mini-games, but also that many challenges remain. For the mini-games we additionally provide scores for two human players: a DeepMind game tester (novice level) and a StarCraft GrandMaster (professional level) (see Table~\ref{tbl:minigames}).

\subsection{Learning Algorithm}

Our reinforcement learning agents are built using a deep neural network with parameters $\theta$, which defines a policy $\pi_{\theta}$.  
At time step $t$ the agent receives observations $s_t$, selects an action $a_t$ with probability $\pi_{\theta}(a_t|s_t)$, and then receives a reward $r_t$ from the environment. The goal of the agent is to maximise the return $G_t = \sum^{\infty}_{k=0}{\gamma^k r_{t+k+1}}$, where $\gamma$ is a discount factor.
For notational clarity we assume that policy is conditioned only on the observation $s_t$, but without loss of generality it might be conditioned on all previous states, e.g., via a hidden memory state as we describe below. 

The parameters of the policy are learnt using Asynchronous Advantage Actor Critic (A3C), as described by Mnih et al. \cite{mnih2016asynchronous}, which was shown to produce state-of-the-art results on a diverse set of environments. A3C is a policy gradient method, which performs an approximate gradient ascent on $\E{G_t}$. The A3C gradient is defined as follows:

\begin{equation}
\underbrace{(G_t - v_{\theta}(s_t))\nabla_{\theta}\log{\pi_{\theta}}(a_t|s_t)}_{\textrm{policy gradient}} + \beta  \underbrace{(G_t - v_{\theta}(s_t)) \nabla_{\theta} v_{\theta}(s_t)}_{\textrm{value estimation gradient}} + \eta \underbrace{\sum_a \pi_{\theta}(a|s_t) \log \pi_{\theta}(a|s_t)}_{\textrm{entropy regularisation}} \,,
\end{equation}

where $v_{\theta}(s)$ is a value function estimate of the expected return $\E{ G_t \mid s_t = s}$ produced by the same network.  Instead of the full return, we use an $n$-step return $G_t = \sum^{n-1}_{k=0}{\gamma^k r_{t+k+1}} ~+~ \gamma^n v_{\theta}(s_{t+n})$ in the gradient above, where $n$ is a hyper-parameter.  The last term regularises the policy towards larger entropy, which promotes exploration, and $\beta$ and $\eta$ are hyper-parameters that trade off the importance of the loss components. For details we refer the reader to the original paper \cite{mnih2016asynchronous} and the references therein.

\subsection{Policy Representation}\label{sec:policy_representation}

As described in section~\ref{sec:env}, the API exposes actions as a nested list $a$ which contains a function identifier $a^0$ and a set of arguments. Since all arguments including pixel coordinates on screen and minimap are discrete, a naive parametrisation of a policy $\pi_{\theta}(a | s)$ would require millions of values to specify the joint distribution over $a$, even for a low spatial resolution.
We could instead represent the policy in an auto-regressive manner, utilising the chain rule\footnote{Note that for the auto-regressive case one could use an arbitrary permutation over arguments to define an order in which the chain rule is applied.  But there is also a `natural' ordering over arguments that can be used since decisions about where to click on a screen depend on the purpose of the click, that is, the identifier of the function being called.}:
\begin{equation}
    \pi_{\theta}(a | s) =
    \prod_{l=0}^L \pi_{\theta}(a^l | a^{<l}, s).
\end{equation}
This representation, if implemented efficiently, is arguably simpler as it transforms the problem of choosing a full action $a$ to a sequence of decisions for each argument $a^l$. 
In the straightforward RL baselines reported here, we make a further simplification and use policies that choose the function identifier, $a^0$, and all the arguments, $a^l$, independently from one another: so, $\pi_{\theta}(a | s) = \prod_{l=0}^L \pi_{\theta}(a^l | s)$. Note that, depending on the function identifier $a^0$, the number of required arguments $L$ is different. Some actions (e.g., the no-op action) do not require any arguments, whereas others (e.g., $\texttt{move\_screen}(x, y)$) do. 
See Figure~\ref{fig:actions} for an example.

In line with the human UI, we ensure that unavailable actions are never chosen by our agents. To do so we mask out the function identifier choice $a^0$ such that only the available subset can be sampled. We implement this by masking out actions and renormalising the probability distribution over $a^0$. 

\subsection{Agent Architectures}

\begin{figure}[ht]
\centering
        \begin{subfigure}[b]{0.46\textwidth}
                \includegraphics[width=\linewidth]{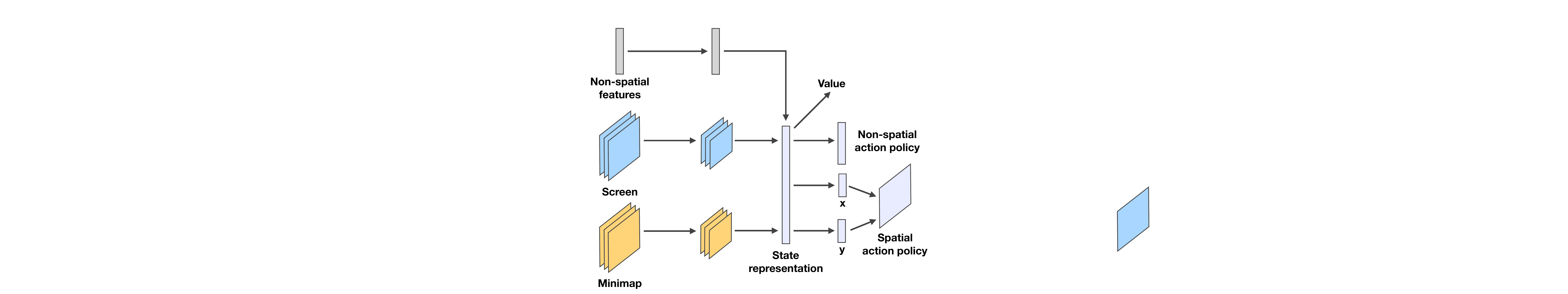}
                \caption{Atari-net}
                \label{fig:atari_net}
        \end{subfigure}%
        \quad \quad
        \begin{subfigure}[b]{0.48\textwidth}
                \includegraphics[width=\linewidth]{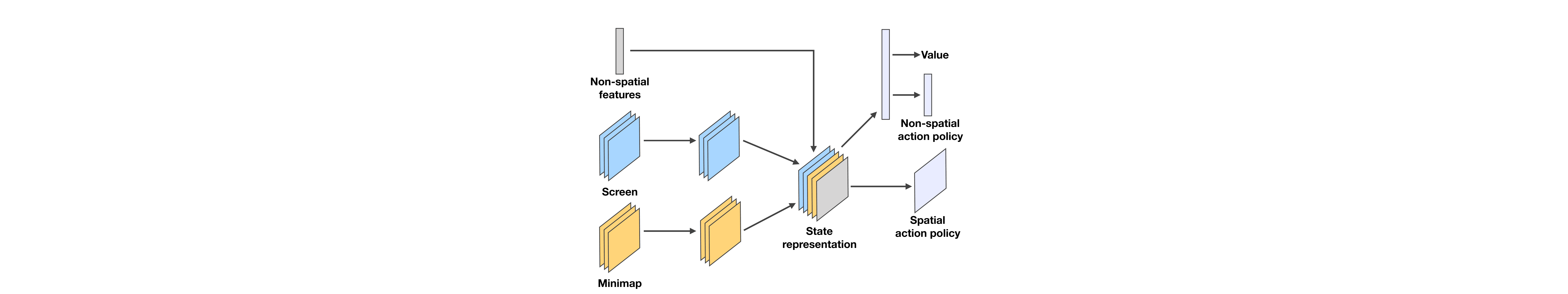}
                \caption{FullyConv}
                \label{fig:conv_net}
        \end{subfigure}%
        \caption{Network architectures of the basic agents considered in the paper.
        }
        \label{fig:networks}
\end{figure}

This section presents several agent architectures with the purpose of producing straightforward baselines. 
We take established architectures from the literature \cite{mnih2015human, mnih2016asynchronous} and adapt them to fit the specifics of the environment, in particular the action space.
Figure~\ref{fig:networks} illustrates the proposed architectures.

\paragraph{Input pre-processing} All the baseline agents share the same pre-processing of input feature layers.
We embed all feature layers containing categorical values into a continuous space, which is equivalent to using a one-hot encoding in the channel dimension followed by a $1 \times 1$ convolution.
We also re-scale numerical features with a logarithmic transformation as some of them such as hit-points or minerals might attain substantially high values.

\paragraph{Atari-net Agent}
The first baseline is a simple adaptation of the architecture successfully used for the Atari~\cite{bellemare2013arcade} benchmark and DeepMind Lab environments~\cite{Beattie:2016}. 
It processes screen and minimap feature layers with the same convolutional network as in~\cite{mnih2016asynchronous} --- two layers with $16,32$ filters of size $8,4$ and stride $4,2$ respectively. 
The non-spatial features vector is processed by a linear layer with a $\textrm{tanh}$ non-linearity. 
The results are concatenated and sent through a linear layer with a $\textrm{ReLU}$ activation. 
The resulting vector is then used as input to linear layers that output policies over the action function id $a^0$ and each action-function argument $\{a^l\}_{l=0}^L$ independently.
For spatial actions (screen or minimap coordinates) we independently model policies to select (discretised) $x$ and $y$ coordinates.

\paragraph{FullyConv Agent} 
Convolutional networks for reinforcement learning (such as the Atari-net baseline above) usually reduce the spatial resolution of the input with each layer and ultimately finish with a fully connected layer that discards spatial structure completely. 
This allows spatial information to be abstracted away before actions are inferred. In StarCraft, though, a major challenge is to infer spatial actions (i.e. clicking on the screen and minimap). 
As these spatial actions act within the same space as the inputs, it might be detrimental to discard the spatial structure of the input.

Here we propose a \textit{fully convolutional} network agent, which predicts spatial actions directly through a sequence of resolution-preserving convolutional layers.
The network we propose has no stride and uses padding at every layer, thereby preserving the resolution of the spatial information in the input. 
For simplicity, we assume the screen and minimap inputs have the same resolution. 
We pass screen and minimap observations through separate 2-layer convolutional networks with $16, 32$ filters of size $5 \times 5, 3 \times 3$ respectively.
The state representation is then formed by the concatenation of the screen and minimap network outputs, as well as the broadcast vector statistics, along the channel dimension.
Note that this is likely non-optimal since the screen and minimap do not have the same spatial extent --- future work could improve on this arrangement.
To compute the baseline and policies over categorical (non-spatial) actions, the state representation is first passed through a fully-connected layer with $256$ units and $\textrm{ReLU}$ activations, followed by fully-connected linear layers.
Finally, a policy over spatial actions is obtained using $1 \times 1$ convolution of the state representation with a single output channel. See Figure~\ref{fig:networks} for a visual representation of this computation.

\paragraph{FullyConv LSTM Agent} 
Both of the above baselines are feed-forward architectures and therefore have no memory. While this is sufficient for some tasks, we cannot expect it to be enough for the full complexity of StarCraft. 
Here we introduce a baseline architecture based on a convolutional LSTM. 
We follow the fully-convolutional agent's pipeline described above and simply add a convolutional LSTM module after the minimap and screen feature channels are concatenated with the non-spatial features.

\paragraph{Random agents}
We use two random baselines. 
\emph{Random policy} is an agent that picks uniformly at random among all valid actions, which highlights the difficulty of stumbling onto a successful episode in a very large action space. 
The \emph{random search} baseline is based on the FullyConv agent and works by taking many independent, randomly initialised policy networks (with a low softmax temperature that induces near-deterministic actions), evaluating each for 20 episodes and keeping the one with the highest mean score. 
This is complementary in that it samples in policy space rather than action space.

\subsection{Results}

In A3C, we truncate the trajectory and run backpropagation after $K=40$ forward steps of a network or if a terminal signal is received. 
The optimisation process runs 64 asynchronous threads using shared RMSProp. For each method, we ran 100 experiments, each using randomly sampled hyper-parameters. Learning rate was sampled from a $\textrm{form}(10^{-5},10^{-3})$ interval. The learning rate was linearly annealed from a sampled value to half the initial rate for all agents. We use an independent entropy penalty of $10^{-3}$ for the action function and each action-function argument. We act at a fix rate every 8 game steps, which is equivalent to about three actions per second or 180 APM. 
All experiments were run for 600M steps (or 8$\times$600M game steps). 

\subsubsection{Full Game}

\begin{figure}[ht]
    \centering
    \includegraphics[width=\linewidth]{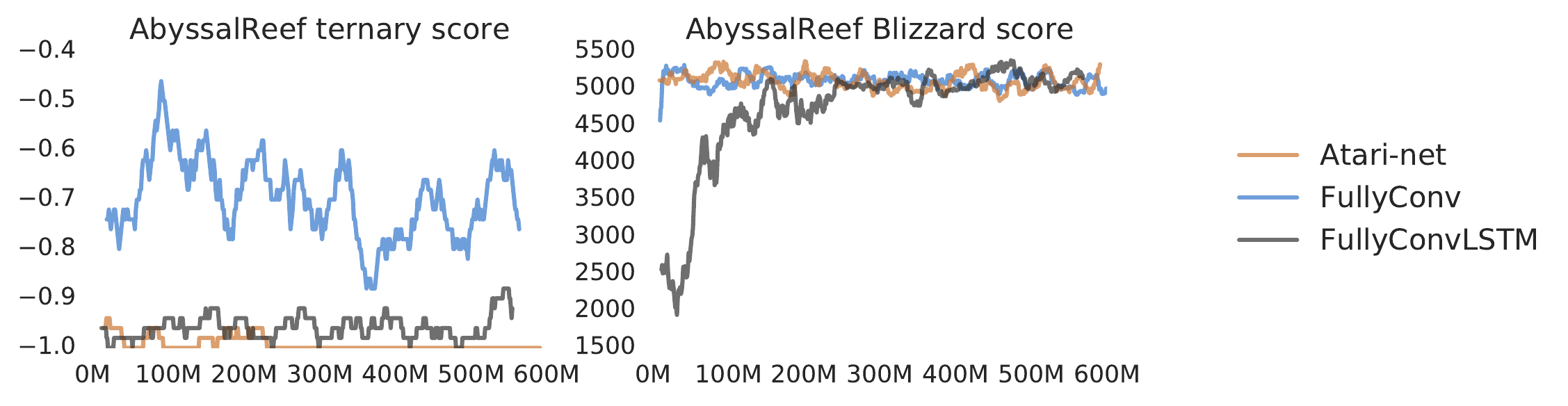}
    \caption{Performance on the full game of the best hyper-parameters versus the easy built-in AI player as the opponent (TvT on the Abyssal Reef LE ladder map): 1. Using outcome (-1 = lose, 0 = tie, 1 = win) as the reward; 2. Using the native game score provided by Blizzard as the reward.  Notably, baseline agents do not learn to win even a single game. Architectures: (a) the original Atari architecture used for DQN, (b) a network which uses a convnet to preserve spatial information for screen and minimap actions, (c) same as in (b) but with a Convolutional LSTM at one layer. Lines are smoothed for visibility.}
    \label{fig:fullgame}
\end{figure}

For experiments on the full game, we selected the Abyssal Reef LE ladder map used in ranked online games as well as in professional matches. 
The agent played against the easiest built-in AI in a Terran versus Terran match-up. Maximum game length was set to 30 minutes, after which a tie was declared, and the episode terminated.  

Results of the experiments are shown on Figure~\ref{fig:fullgame}. 
Unsurprisingly, none of the agents trained with sparse ternary rewards developed a viable strategy for the full game.
The most successful agent, based on the fully convolutional architecture without memory, managed to avoid constant losses by using the Terran ability to lift and move buildings out of attack range.  This makes it difficult for the easy AI to win within the 30 minute time limit.

Agents trained with the Blizzard score converged to trivial strategies that avoid distracting workers from mining minerals. 
Most agents converged to simply preserving the initial mining process without building further units or structures (this behaviour was also observed in the economic mini-game proposed below).

These results suggest that the full game of StarCraft II is indeed a challenging RL domain, especially without access to other sources of information such as human replays.

\subsubsection{Mini-Games}

\begin{figure}[ht!]
    \centering
    \includegraphics[width=\linewidth]{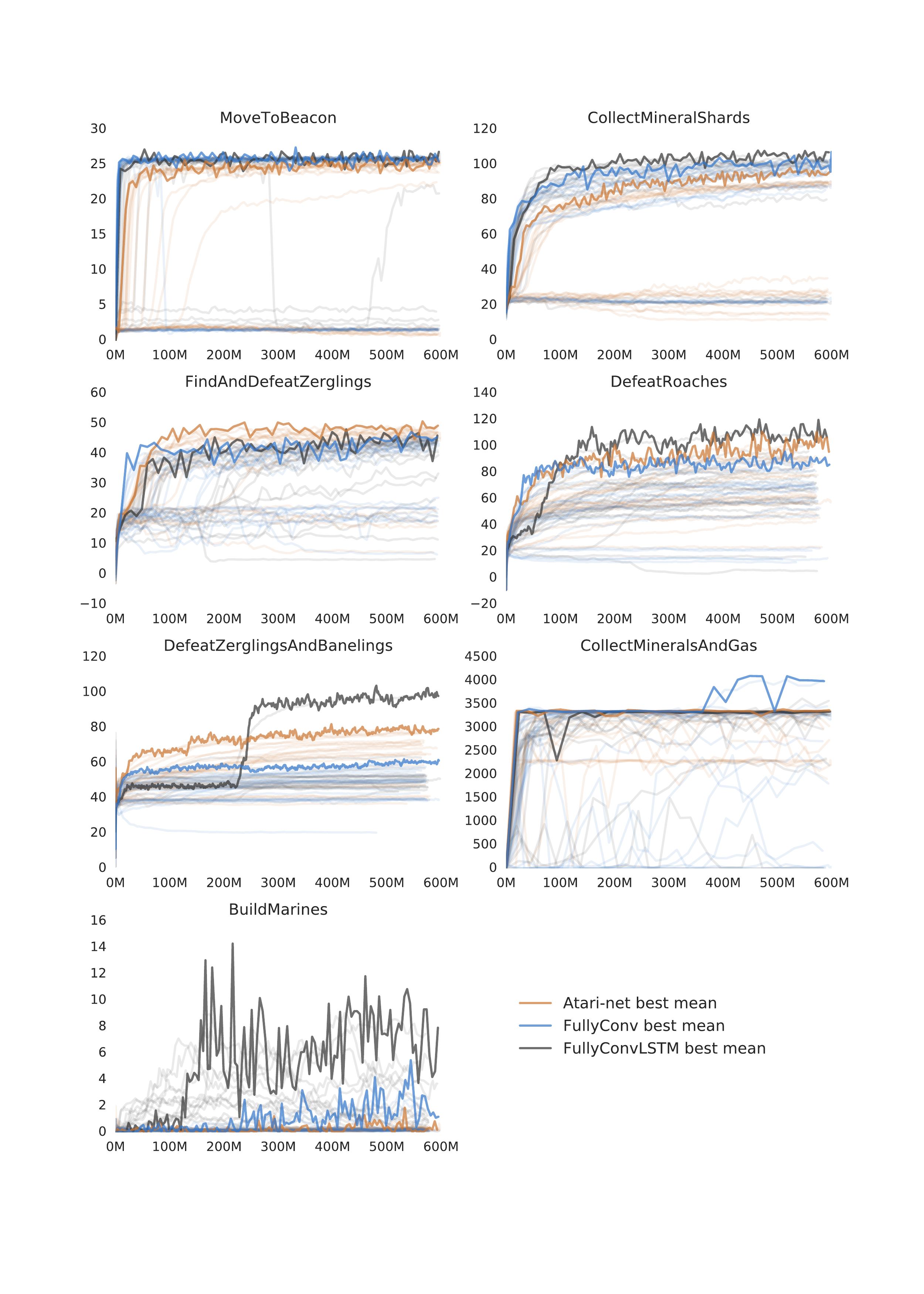}
    \caption{Training process for baseline agent architectures. 
    Displayed lines are mean scores as a function of game steps. 
    The three network architectures are the same as used in Figure~\ref{fig:fullgame}. Faint lines show all 100 runs with different hyper-parameters; the solid line is the run with the best mean.  Lines are smoothed for visibility.
    }
    \label{fig:minigames}
\end{figure}

As described in section~\ref{sec:env}, one can avoid the complexity of the full game by defining a set of mini-games which focus on certain aspects of the game (see section~\ref{sec:env} for a high-level description of each mini-game). 

\begin{table}[]
~
\centering
\caption{Aggregated results for human baselines and agents on mini-games. All agents were trained for 600M steps.
\textsc{Mean} corresponds to the average agent performance, 
\textsc{best mean} is the average performance of the best agent across different hyper-parameters, \textsc{max} corresponds to the maximum observed individual episode score.
}
\label{tbl:minigames}
\begin{tabular}{c|c|rrrrrrr}
    \sc \textbf{Agent}
    & \sc \textbf{Metric} 
    & \multicolumn{1}{c}{\sc \rotatebox[origin=l]{90}{MoveToBeacon}}
    & \multicolumn{1}{c}{\sc \rotatebox[origin=l]{90}{CollectMineralShards}}
    & \multicolumn{1}{c}{\sc \rotatebox[origin=l]{90}{FindAndDefeatZerglings}}
    & \multicolumn{1}{c}{\sc \rotatebox[origin=l]{90}{DefeatRoaches}}
    & \multicolumn{1}{c}{\sc \rotatebox[origin=l]{90}{DefeatZerglingsAndBanelings}}
    & \multicolumn{1}{c}{\sc \rotatebox[origin=l]{90}{CollectMineralsAndGas}}
    & \multicolumn{1}{c}{\sc \rotatebox[origin=l]{90}{BuildMarines}}
    \\ \hline
\multirow{2}{*}{\sc Random policy}
    & \sc mean      & 1    & 17 & 4    & 1    & 23    & 12      & $<1$  \\
    & \sc max       & 6    & 35 & 19   & 46   & 118   & 750     & 5  \\ \hline
\multirow{2}{*}{\sc Random search}   
    & \sc mean & 25   & 32 & 21   & 51   & 55    & 2318    & 8   \\
    & \sc max       & 29   & 57 & 33   & 241  & 159   & 3940    & 46  \\ \hline
\multirow{2}{*}{\sc DeepMind human player}
    & \sc mean      & 26   & 133 & 46   & 41   & 729   & 6880    & 138  \\
    & \sc max       & 28   & 142 & 49   & 81   & 757   & 6952    & 142  \\ \hline
\multirow{2}{*}{\sc StarCraft GrandMaster}
    & \sc mean      & 28   & 177 & 61   & 215  & 727   & 7566    & 133   \\
    & \sc max       & 28   & 179 & 61   & 363  & 848   & 7566    & 133   \\ \hline\hline
\multirow{2}{*}{\sc Atari-net}

    & \sc best mean &  25   & 96 & 49   & 101  & 81    & 3356    & $<1$ \\
    & \sc max       & 33   & 131 & 59   & 351  & 352   & 3505    & 20  \\ \hline
\multirow{2}{*}{\sc FullyConv}
    & \sc best mean & 26   & 103 & 45   & 100  & 62    & 3978    & 3   \\
    & \sc max       & 45   & 134 & 56   & 355  & 251   & 4130    & 42  \\ \hline
\multirow{2}{*}{\sc FullyConv LSTM}
    & \sc best mean & 26   & 104 & 44   & 98   & 96    & 3351    & 6   \\
    & \sc max       & 35   & 137 & 57   & 373  & 444   & 3995    & 62
\end{tabular}
\end{table}

We trained our agents on each mini-game. 
The aggregated training results are shown in Figure~\ref{fig:minigames} and the final results with comparisons to human baselines can be found in Table~\ref{tbl:minigames}. A video showcasing our agents can also be found at \url{https://youtu.be/6L448yg0Sm0}.

Overall, fully convolutional agents performed the best across the non-human baselines.
Somewhat surprisingly, the Atari-net agent appeared to be quite a strong competitor on mini-games involving combat, namely FindAndDefeatZerlings, DefeatRoaches and DefeatZerlingsAndBanelings.
On CollectMineralsAndGas, only the best Convolutional agent learned to increase the initial resource income by producing more worker units and assigning them to mining.

We found BuildMarines to be the most strategically demanding mini-game and perhaps the closest of all to the full game of StarCraft. The best results on this game were achieved by FullyConv LSTM and Random Search, while Atari-Net failed to learn a strategy to consistently produce marines during each episode.
It should be noted that, without the restrictions on action space imposed by this map, it would be significantly more diffucult to learn a to produce marines in this mini-game.

All agents performed sub-optimally when compared against the GrandMaster player, except for in simplest MoveToBeacon mini-game, which only requires good mechanics and reaction time --- which artificial agents are expected to be good at. However, in some games like DefeatRoaches and FindAndDefeatZerglings, our agents did fare well versus the DeepMind game tester.

The results of our baseline agents demonstrate that even relatively simple mini-games present interesting challenges for existing RL algorithms.
\section{Supervised Learning from Replays}\label{sec:sl}

Game replays are a crucial resource used by professional and amateur players alike, who learn new strategies, find critical mistakes made in a game, or simply enjoy watching others play as a form of entertainment. Replays are especially important in StarCraft because of hidden information:  the fog-of-war hides all of the opponent's units unless they are within view of one of your own. Thus, among professional players it is standard practice to review and analyse every game they play, even when they win. 

The use of supervised data such as replays or human demonstrations has been successful in robotics \cite{argall2009survey,popov2017data}, the game of Go \cite{maddison2014move,silver2016mastering}, and Atari \cite{hester2017learning}. It has also been used in the context of StarCraft I (e.g., \cite{justesen2017learning}), though not to train a policy over basic actions, but rather to discover build orders. StarCraft II provides the opportunity to collect and learn from a large and growing set of human replays. Whereas there has been no central and standardised mechanism for collecting replays for StarCraft I, large numbers of anonymised StarCraft II games are readily available via Blizzard's online 1v1 ladder.  As well, more games will be added to this set on a regular basis as a relatively stable player pool plays new games.

Learning from replays should be useful to bootstrap or complement reinforcement learning.  In isolation, it could also serve as a benchmark for sequence modelling or memory architectures having to deal with long term correlations. Indeed, to understand a game as it unfolds, one must integrate information across many time steps efficiently. Furthermore, due to partial observability, replays could also be used to study models of uncertainty such as (but not limited to) variational autoencoders \cite{Kingma:2014}. Finally, comparing performance on outcome/action prediction may help guide the design of neural architectures with suitable inductive biases for RL in the domain.

In the rest of this section, we provide baselines using the architectures described in Section~\ref{sec:rl}, but using a set of 800K games to learn both a value function (i.e., predicting the winner of the game from game observations), and a policy (i.e., predicting the action taken from game observations). The games contain all possible matchups in StarCraft II (i.e., we do not restrict the agent to play a single race). 

\begin{figure}[htp]
    \centering
    \begin{subfigure}{.5\textwidth}
      \centering
      \includegraphics[width=.95\linewidth]{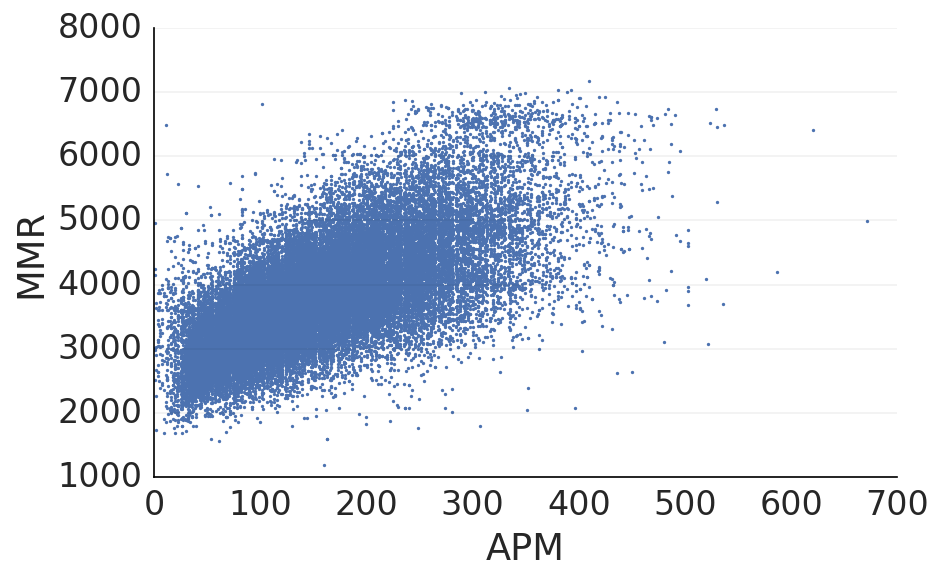}
    \end{subfigure}%
    \begin{subfigure}{.5\textwidth}
      \centering
      \includegraphics[width=.95\linewidth]{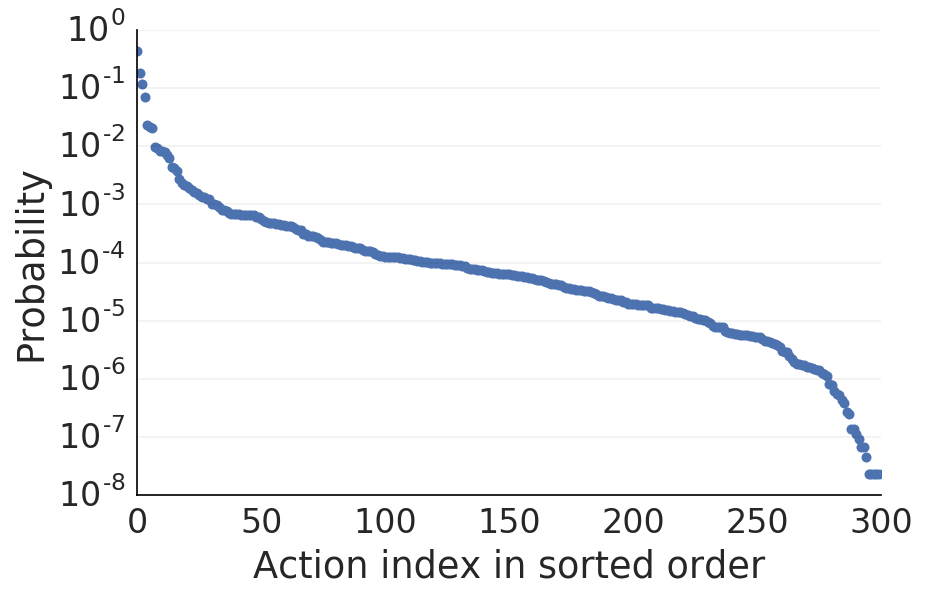}
    \end{subfigure}
    \caption{Statistics of the replay set we used for supervised training of our policy and value nets.
    (Left) Distribution of player rating (MMR) as a function of APM.
    (Right) Distribution of actions sorted by probability of usage by human players.
    }
    \label{fig:replay_stats}
\end{figure}

Figure~\ref{fig:replay_stats} shows statistics for the replays we used. We summarize some of the most interesting stats here: 1. The skill level of players, measured by the Match Making Rating (MMR), varies from casual gamer, to high-end amateur, on through to professionals. 2. The average number of Actions Per Minute (APM) is $153$, and mean MMR is $3789$.  3. The replays are not filtered, and instead all `ranked' league games played on BattleNet are used \footnote{\url{http://wiki.teamliquid.net/starcraft2/Battle.net\_Leagues}}.
4. Less than one percent are Masters level replays from top players. 5. We also show the distribution of actions sorted by their frequency of use by human players. The most frequent action, taken $43\%$ of the time, is moving the camera. 6. Overall, the action distribution has a heavy tail with a few commonly used actions (e.g., move camera, select rectangle,  attack screen) and a large number of actions that are used infrequently (e.g., building an engineering bay).

We train dual-headed networks that predict both the game outcome ($1=\,$win vs. $0=\,$loss or tie), and the action taken by the player at each time step. Sharing the body of the network makes it necessary to balance the weights for the two loss functions, but it also allows value and policy predictions to inform one another. We did not make ties a separate game outcome class in the supervised training setup, since the number of ties in the dataset is very low ($< 1$\%) compared to victory and defeat

\subsection{Value Predictions}

Predicting the outcome of a game is a challenging task. Even professional StarCraft II commentators often fail to predict the winner despite having a full access to the game state (i.e., not being limited by partial observability). Value functions that accurately predict game outcomes are desirable because they can be used to alleviate the challenge of learning from sparse rewards. From given state, a well trained value function can suggest which neighbouring states would be worth moving into long before seeing the game outcome.

Our setup for supervised learning begins with the straightforward baseline architectures described in Section~\ref{sec:rl}: Atari-net and FullyConv. The networks do not take into account previous observations, i.e., they predict the outcome from a single frame (this is clearly sub-optimal). Furthermore, the observation does not include any privileged information: an agent has to produce value predictions based only on what it can see at any given time step (i.e. fog-of-war is enabled).  Thus, if the opponent has managed to secretly produce many units that are very effective against the army that the agent has built, it may mistakenly believe that its position is stronger than it is.

\begin{figure}[htp]
    \centering
    \begin{subfigure}{.5\textwidth}
      \centering
      \includegraphics[width=.95\linewidth]{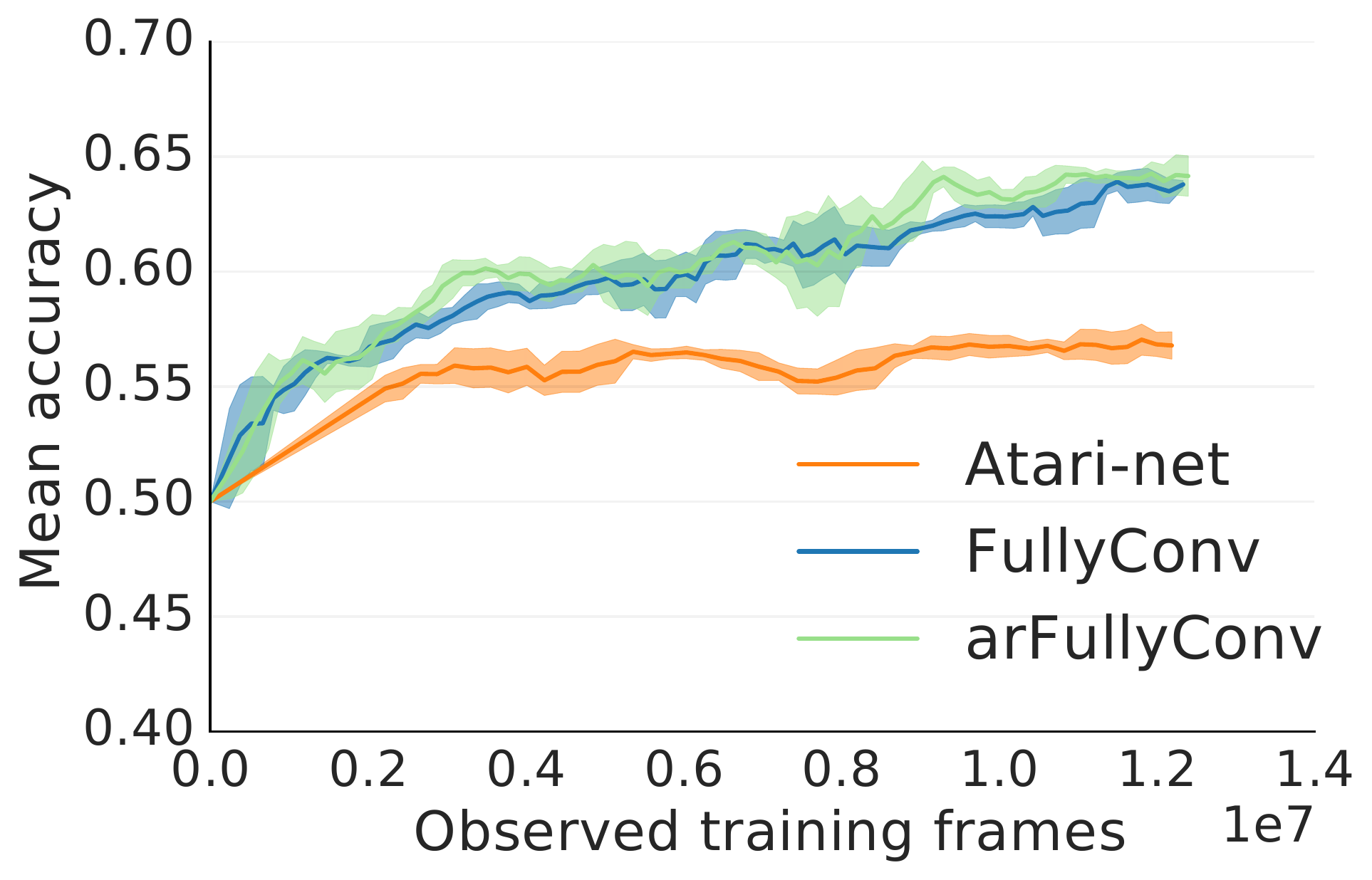}
    \end{subfigure}%
    \begin{subfigure}{.5\textwidth}
      \centering
      \includegraphics[width=.95\linewidth]{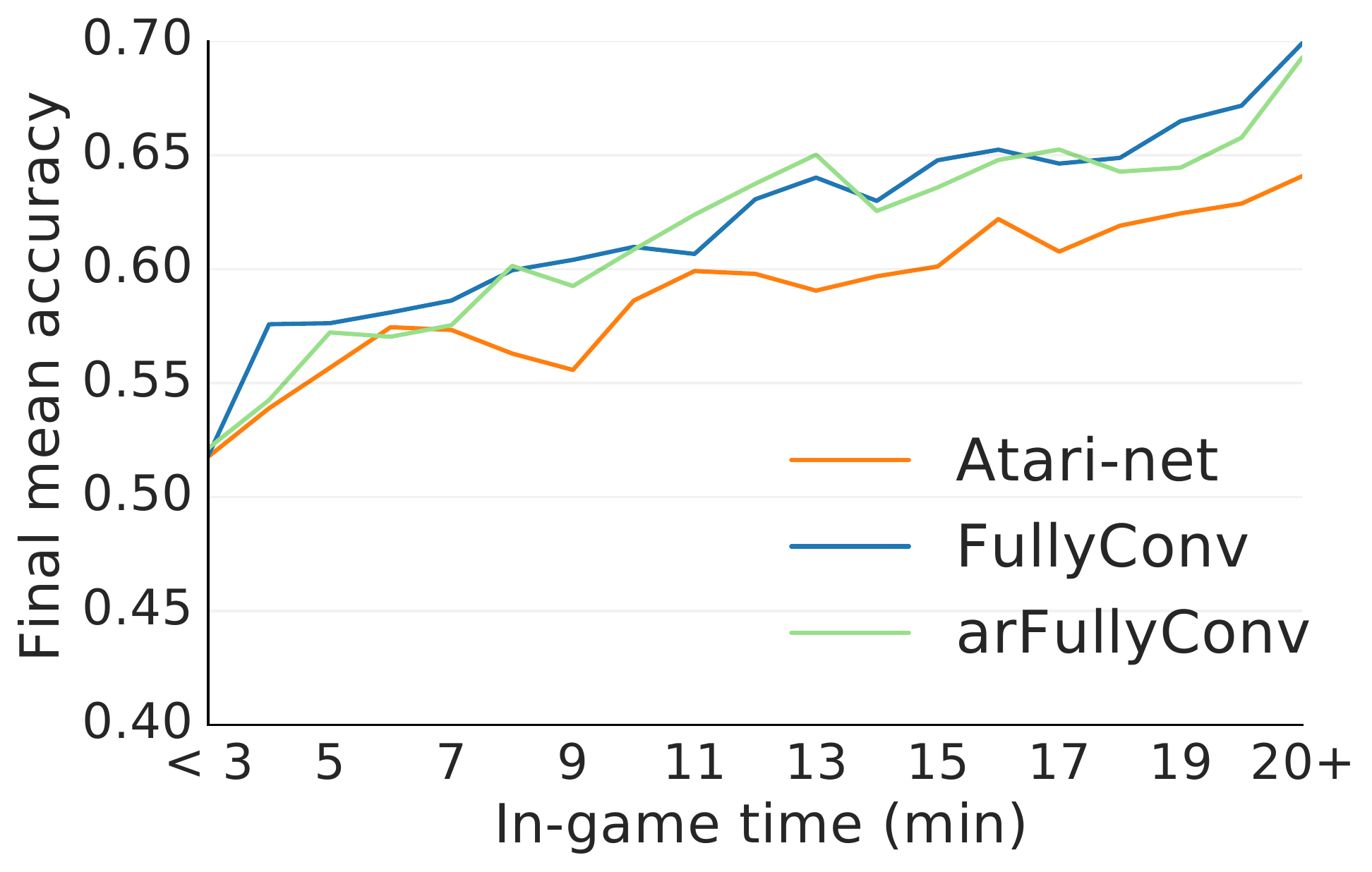}
    \end{subfigure}
    \caption{The accuracy of predicting the outcome of StarCraft games using a network that operates on the screen and minimap feature planes as well as the scalar player stats. 
    (Left) Train curves for three different network architectures.
    (Right) Accuracy over game time. At the beginning of the game (before 2 minutes), the network has 50\% accuracy (equivalent to chance). This is expected since the outcome is less clear earlier in the game. By the 15 minute mark, the network is able to correctly predict the winner 65\% of the time.}
    \label{fig:value}
\end{figure}

The networks proposed in Section~\ref{sec:rl} produce the action identifier and its arguments independently. However, the accuracy of predicting a point on the screen can be improved by conditioning on the base action, e.g., building an extra base versus moving an army. Thus, in addition to the Atari-net and FullyConv architecture, we have arFullyConv which uses the auto-regressive policy introduction introduced in Section~\ref{sec:policy_representation}, i.e. using the function identifier $a^0$ and previously sampled arguments $a^{<l}$ to model a policy over the current argument $a^l$.

Networks are trained for $200$k steps of gradient descent on all possible match-ups in StarCraft II.
We trained with mini-batches of $64$ observations taken at random from all replays uniformly across time. Observations are sampled with a step multiplier of $8$, consistent with the RL setup. The resolution of both screen and minimap is $64\times 64$. Each observation consists of the screen and minimap spatial feature layers as well as player stats such as food cap and number of collected minerals that human players see on the screen.
We use $90$\% of the replays as training set, and a fixed test set of $0.5$M frames drawn from the rest of the $10$\% of the replays. The agent performance is evaluated continuously against this test set as training progresses.  

Figure~\ref{fig:value} shows average accuracy over training step as well as accuracy of a trained model as a function of game time.
%\karen{TODO: The Figure refers to arConvolutional model which is undefined at this point.}
A random baseline would correct approximately 50\% of the time since the game is well balanced across all race pairs, and tying is extremely rare. As training progresses, the FullyConv architecture achieves an accuracy of 64\%.  Also, as the game progresses, value prediction becomes more accurate, as seen in Figure~\ref{fig:value} (Right). This mirrors the results of prior work on StarCraft~I~\cite{erickson2014global}.

\subsection{Policy Predictions}

\begin{table}[htp]
\centering
 \begin{tabular}{ r | r r r | r r r } 
% & \multicolumn{3}{c|}{cross entropy} & \multicolumn{3}{c|}{top 1 accuracy} & \multicolumn{3}{c|}{top 5 accuracy}  \\
 & \multicolumn{3}{c|}{\sc top 1 accuracy} & \multicolumn{3}{c}{
 \sc top 5 accuracy}  \\
 \cline{2-7}
 %& action & screen & minimap & action & screen & minimap & action & screen & minimap  \\ 
  & \sc action & \sc screen & \sc minimap & \sc action & \sc screen & \sc minimap  \\ 
 \hline
%  \sc ATARI & 1.836	& 7.406	& 3.670	& 0.378	& 0.012	& 0.198 & 0.872 & 0.029 & 0.556 \\
%  CNN & 1.831 & 6.685 &	3.228 & 0.379 &	0.095 &	0.257 & 0.882 &	0.185 &	0.623 \\
%  aCNN & 1.842 &	6.359 &	3.091 &	0.377 &	0.105 &	0.259 &	0.874 &	0.221 &	0.627 \\ 
%  Random & 3.021 &	8.589 &	8.543 &	0.080 &	0.000 & 0.000 & 0.378 &	0.001 &	0.001 \\
  \sc Atari-net & 37.8\%	& 1.2\%	& 19.8\% & 87.2\% & 2.9\% & 55.6\% \\
 \sc FullyConv & 37.9\% &	9.5\% &	25.7\% & 88.2\% &	18.5\% &	62.3\% \\
 \sc arFullyConv & 37.7\% &	10.5\% &	25.9\% &	87.4\% &	22.1\% &	62.7\% \\ 
 \hline
 \sc Random & 4.3\% &	0.0\% & 0.0\% & 29.5\% &	1.0\% &	1.0\% \\
\end{tabular}
\caption{Policy top 1 and top 5 accuracies for the base actions and screen/minimap arguments. 
arFullyConv refers to the autoregressive version of  FullyConv. The random baseline is a arFullyConv with randomly initialised weights.}\label{tab:policy}
\end{table}

The same network trained to predict values had a separate output designed to predict the action issued by the user. We sometimes refer to this part of the network as the policy since it can be readily deployed to play the game. 

There are many schemes one might employ to train networks to imitate human behaviour from replays.  Here we use a simple approach that connects straightforwardly with the RL work in Section~\ref{sec:rl}. When training our policy we sampled observations at a fixed step multiplier of $8$ frames.  
We take the first action issued within each $8$ frames as the learning target for the policy.
If no action was taken during that period, we take the target to be a `no-op', i.e., a special action which has no effect. 

When humans play StarCraft II, only a subset of all possible actions are available at any given time. For example, ``building a marine'' is enabled only if barracks are currently selected. Networks should not need to learn to avoid illegal actions since this information is readily available.  Thus, during training, we filter out actions that would not be available to a human player. To do so, we take the union of all available actions for the past $8$ frames and apply a mask that sets the probability of all unavailable actions to near zero.

Note that, as previously mentioned, we trained the policy to play all possible matchups. Thus, in principle, the agent can play any race. However, for consistency with the reinforcement learning agents studied in Section~\ref{sec:rl}, we report in-game metrics in the single Terran versus Terran matchup.

Table~\ref{tab:policy} shows how different architectures perform in terms of accuracy at predicting the action identifier, the screen, and the minimap argument. As expected, both FullyConv and arFullyConv architectures perform much better for spatial arguments.  As well, the arFullyConv architecture outperforms FullyConv, presumably because it knows which action identifier the argument will be used for.

\begin{figure}[htp]
    \centering
    \begin{subfigure}{.5\textwidth}
      \centering
      \includegraphics[width=.95\linewidth]{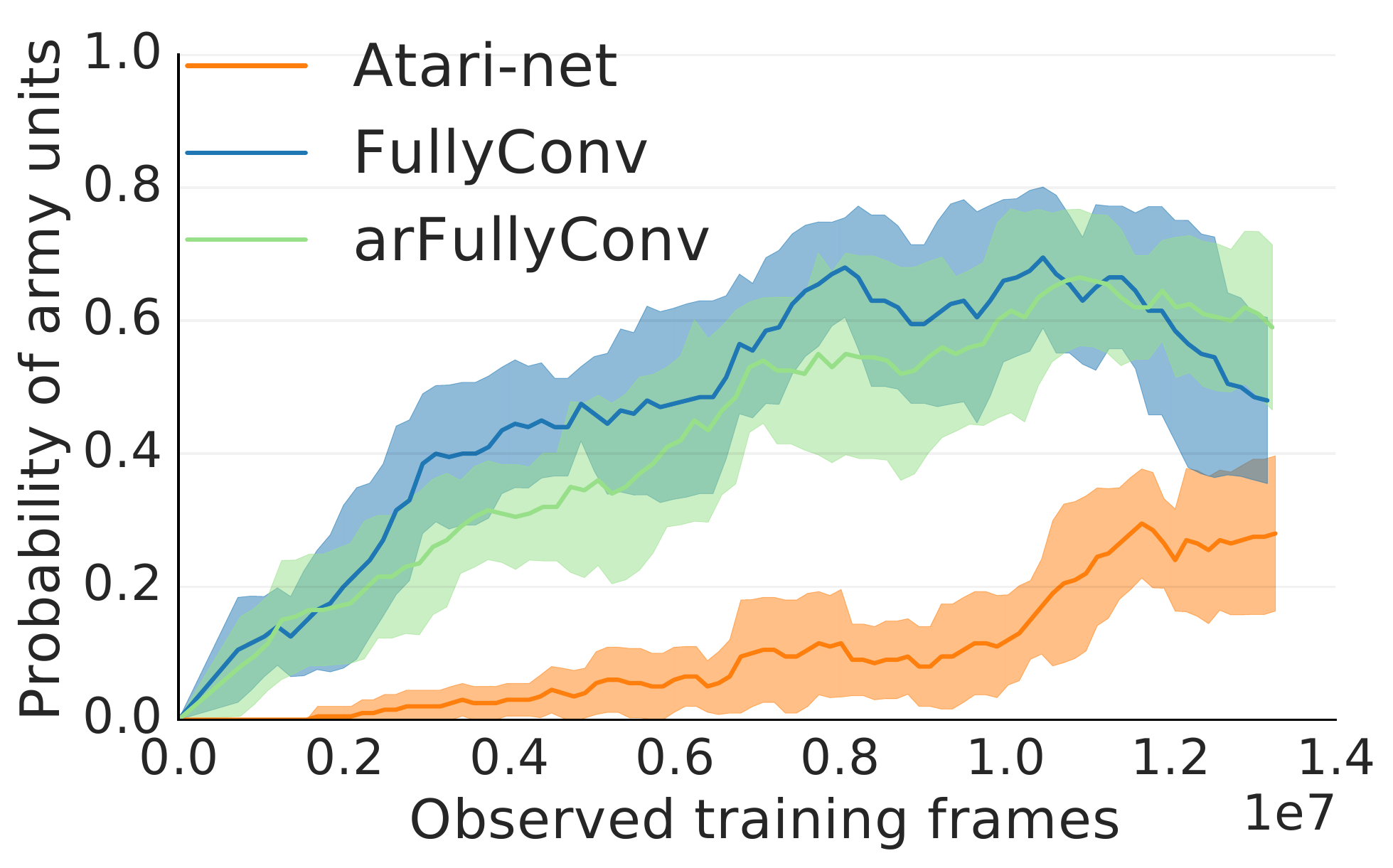}
    \end{subfigure}%
    \begin{subfigure}{.5\textwidth}
      \centering
      \includegraphics[width=.95\linewidth]{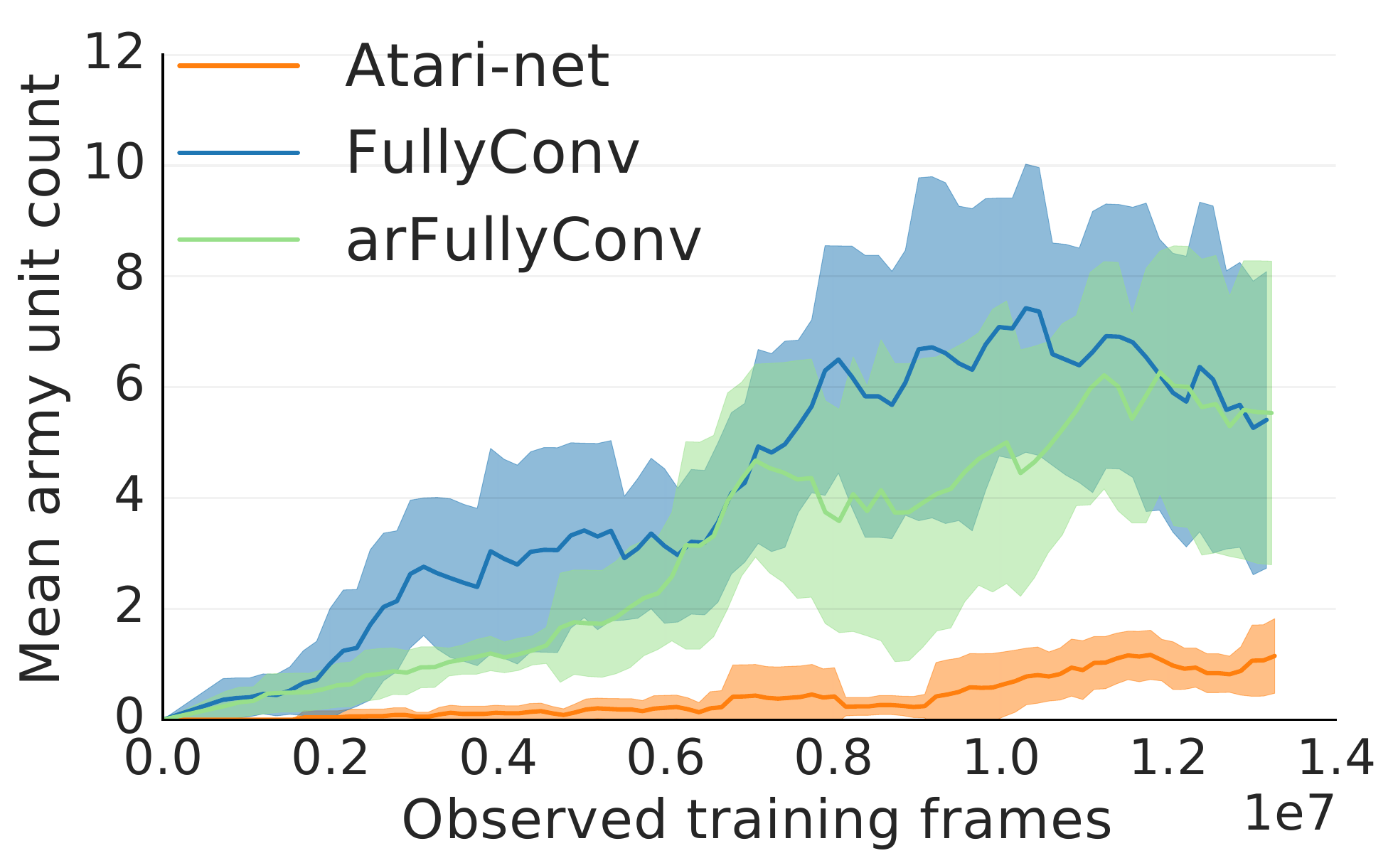}
    \end{subfigure}
    \caption{The probability of building army units as training the policy nets progresses over the training data. The game setup is Terran vs. Terran.
    (Left) Probability of building any army units in a game.
    (Right) Average number of army units built per game.}
    \label{fig:policy}
\end{figure}

When we directly plug the policy trained with supervised learning into the game, it is able to produce more units and play better as a function of observed replay data, as shown in Figure~\ref{fig:policy} and in the video at \url{https://youtu.be/WEOzide5XFc}. It also outperforms all agents trained in Section~\ref{sec:rl} on the simpler  mini-game of BuildMarines, which has a restricted action space, even though the supervised policy is playing an unrestricted, full 1v1 game. These results suggest that supervised imitation learning may be a promising direction for bootstrapping StarCraft II agents. Future work should look to improve imitation initialised policies by training directly with reinforcement learning on the objective we really care about -- i.e., the game outcome.

\section{Conclusions \& Future Work}

This paper introduces StarCraft II as a new challenge for deep reinforcement learning research.  We provide details for a freely available Python interface to play the game as well as human replay data from ranked games collected via Blizzard's official BattleNet ladder.  With this initial release we describe supervised learning results on the human replay data for policy and value networks.  We also also describe results for straightforward baseline RL agents on seven mini-games and on the full game.  

We regard the mini-games primarily as unit tests.  That is, an RL agent should be able to achieve human level performance on these with relative ease if it is to have a chance to succeed on the full game.  It may be instructive to build additional mini-games, but we take the full game --- evaluated on the final outcome --- as the most interesting problem, and hope first and foremost to encourage research that will lead to its solution.

While performance on some mini-games is close to expert human play, we find, as expected, that current state-of-the-art baseline agents cannot learn to win against the easiest built-in AI on the full game.  This is true not only when the game outcome (i.e., -1, 0, 1) is used as the reward signal, but also when a shaping reward is provided at each timestep (i.e., the native game score provided by Blizzard).
In this sense, our provided environment presents a challenge that is at once canonical, externally defined, and completely intractable for off-the-shelf baseline algorithms.

This release simplifies several aspects of the game as it is played by humans: 1. the observations are preprocessed before they are given to the agent, 2. the action space has been simplified to be more easily used by RL agents instead of using the keyboard and mouse-click setup used by humans, 3. it is played in lock-step so that agents can compute for as long as they need at each time-step rather than being real-time, and 4. the full game only allows play against the built-in AI. However, we consider the \emph{real} challenge to build agents that can play the best human players on their own turf, that is with RGB pixel observations and strict time limits. Therefore, future releases may relax the simplifications above, as well as enable self-play, moving us towards the goal of training agents that humans consider to be fair opponents.

\section*{Contributions}

Blizzard:

\begin{itemize}
    \item StarCraft II Binary
    \item StarCraft II API: \url{https://github.com/Blizzard/s2client-proto}
    \item Replays
\end{itemize}

DeepMind:

\begin{itemize}
    \item PySC2: \url{https://github.com/deepmind/pysc2}
    \item All the agents and experiments in the paper
\end{itemize}

\section*{Acknowledgements}

We would like to thank many at Blizzard, especially Tommy Tran, Tyler Plass, Brian Song, Tom van Dijck, and Greg Risselada, the Grandmaster. We would also like to thank the DeepMind team, especially Nal Kalchbrenner, Ali Eslami, Jamey Stevenson, Adam Cain and our esteemed game testers Amir Sadik \& Sarah York. We also would like to thank David Churchill for his early feedback on the Raw API, for building CommandCenter, and for comments on the manuscript.

\bibliographystyle{plainnat}
\bibliography{references}
\end{document}